\documentclass{bmvc2k}

\usepackage{booktabs}
\usepackage{multirow}
\usepackage{amssymb}
\usepackage{float}
\usepackage{caption}
\newcommand{\teach}{$\mathcal{T}$ }
\newcommand{\stud}{$\mathcal{S}$ }

\title{SimReg: Regression as a Simple Yet Effective Tool for Self-supervised Knowledge Distillation}

\addauthor{K L Navaneet}{navanek1@umbc.edu}{1}
\addauthor{Soroush Abbasi Koohpayegani}{soroush@umbc.edu}{1}
\addauthor{Ajinkya Tejankar}{at6@umbc.edu}{1}
\addauthortwo{Hamed Pirsiavash}{hpirsiav@ucdavis.edu}{1}{2}

\addinstitution{
 University of Maryland, Baltimore County\\
 Maryland, USA
}
\addinstitution{
 University of California, Davis \\
 California, USA
}

\runninghead{Navaneet, Koohpayegani, Tejankar, Pirsiavash}{Regressed SSL distillation}

\def\etal{\emph{et al}\bmvaOneDot}

\begin{document}

\maketitle

\begin{abstract}
Feature regression is a simple way to distill large neural network models to smaller ones. We show that with simple changes to the network architecture, regression can outperform more complex state-of-the-art approaches for knowledge distillation from self-supervised models. Surprisingly, the addition of a multi-layer perceptron head to the CNN backbone is beneficial even if used only during distillation and discarded in the downstream task. Deeper non-linear projections can thus be used to accurately mimic the teacher without changing inference architecture and time. Moreover, we utilize independent projection heads to simultaneously distill multiple teacher networks. We also find that using the same weakly augmented image as input for both teacher and student networks aids distillation. Experiments on ImageNet dataset demonstrate the efficacy of the proposed changes in various self-supervised distillation settings. Code is available at \url{https://github.com/UCDvision/simreg}
\end{abstract}

\section{Introduction}
\label{sec:intro}

There has been a tremendous improvement in deep learning methodologies and architectures in the last few years. 
While this has lead to significant improvements in performance on various computer vision tasks, it has also
resulted in complex and deep networks that require high compute during 
inference~\cite{he2016deep,huang2017densely,huang2019gpipe,zagoruyko2016wide}. Various specialized 
architectures~\cite{tan2019efficientnet,iandola2016squeezenet,howard2017mobilenets,sandler2018mobilenetv2} have 
been proposed to minimize the inference time and memory requirements of the model to be deployed. 
Knowledge distillation~\cite{bucilua2006model,hinton2015distilling} has been proposed as an effective technique to
compress information from larger but effective models (teachers) to lighter ones (students). 

With availability of large scale unlabeled datasets, self-supervised
learning (SSL) has received great attention in recent times. Several SSL methods achieve close to
supervised performance on the benchmark ImageNet object classification task~\cite{caron2018deep, grill2020bootstrap}. 
Unlike supervised models, the outputs of a self-supervised network are latent feature vectors 
and not class probabilities. An additional module is generally trained atop the pretrained SSL models using 
supervision to perform the downstream task. Conventional 
knowledge distillation methods proposed for supervised classification are thus not applicable for distillation from 
self-supervised networks. A simple way to handle this is to directly regress the teacher latent features. 
Recent works~\cite{koohpayegani2020compress,fang2021seed}
have proposed more complex solutions that try to capture the structure of the teacher latent space and are shown to
outperform the regression baselines.

\begin{figure}
    \centering
    \includegraphics[width=0.9\linewidth]{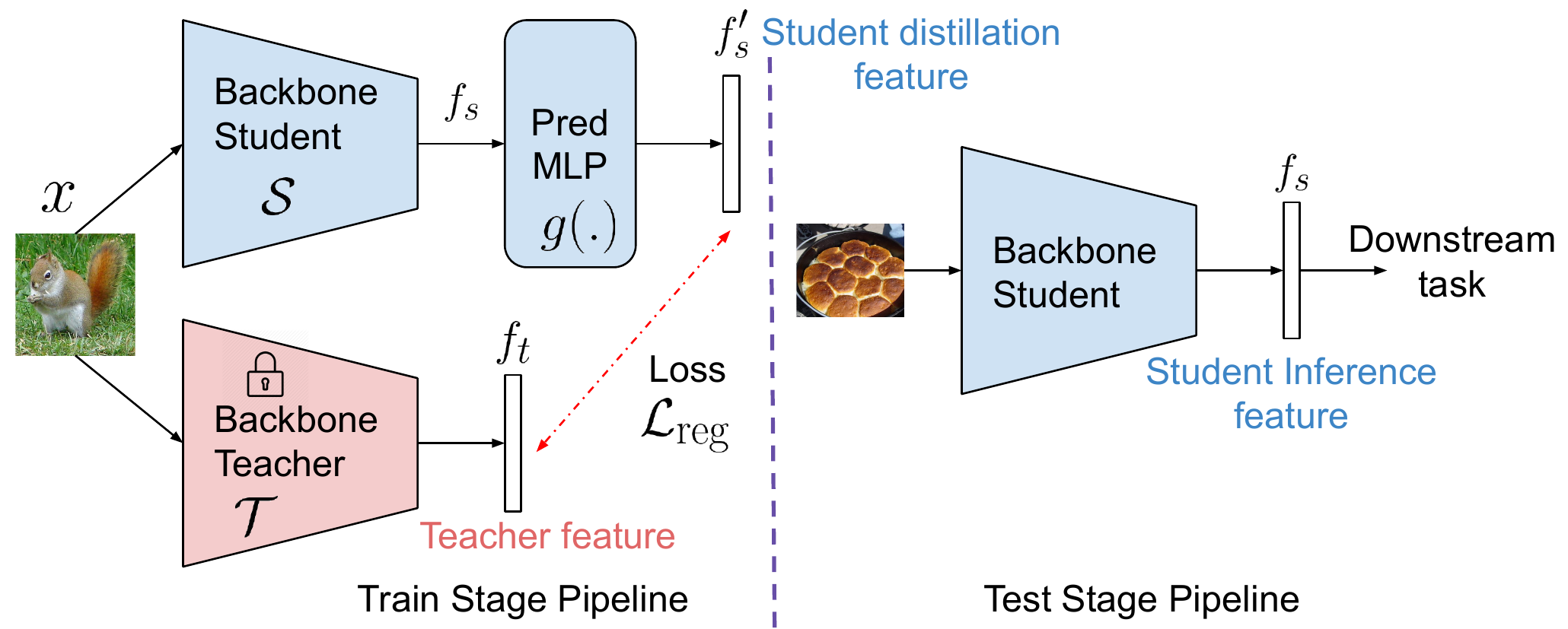}
    \caption{\textbf{Proposed distillation pipeline: }We propose a simple modification of using a MLP prediction module during distillation. The module is discarded during inference. Surprisingly, performance of backbone features $f_s$ is better than those from MLP output, $f_s'$, though $f_s'$ more closely matches the teacher. A deeper MLP helps improve distillation performance.}
    \label{fig:arch_reg}
\end{figure}

Use of a multi-layer perceptron (MLP) head atop CNN backbone model has been shown to help self-supervised models
prevent overfitting to the SSL task and generalize better to downstream 
applications~\cite{chen2020simple,chen2020improved,grill2020bootstrap,chen2021exploring}.
Such modules are used only during SSL pre-training and are not part of inference network.
In this work, we consider the task of distilling self-supervised models. 
We employ a similar \textit{prediction head} atop the student backbone network to effectively mimic the teacher.
As in SSL, the prediction module is discarded after distillation and
thus, there is no change in the time and memory required during inference (refer Fig.~\ref{fig:arch_reg}). 
We empirically demonstrate that doing so does not hurt classification
performance. Counter-intuitively, we observe that the features from the backbone network outperform those from 
the final layer of the prediction head though the final layer best matches the teacher.  
Unlike in SSL, 
overfitting to the training task (i.e, exactly mimicking the teacher) benefits 
distillation~\cite{beyer2021knowledge} and it is not clear why generalization 
could be better at intermediate layers where the similarity with teacher is reduced. 
Our finding suggests that we require a deeper analysis 
to understand how well the student models mimic the teacher in general and how knowledge distillation works. 
Crucially, it also enables us to use a deeper prediction head to achieve lower train and test error
leading to better downstream performance without increasing the student capacity.

We empirically show that the above observation generalises to distillation with different teacher and student settings 
and to other self-supervised distillation technique.
Our simple regression model with a MLP prediction head 
outperforms complex state-of-the-art approaches that require the use of memory banks and tuning of temperature parameter.
Our work serves as an important benchmark for future self-supervised distillation works.
The use of MLP heads also facilitates effective distillation from multiple SSL teacher networks.
Additionally, we demonstrate that using the same augmented image with weak augmentation for both student
and teacher networks results in better student models. Since aggressive augmentation is necessary for effective
self-supervised learning~\cite{chen2020simple,chen2020improved} but hurts their ability to
generalize~\cite{purushwalkam2020demystifying}, our approach could be used to learn better SSL models.
To summarize, our contributions are simple changes to architecture and augmentation strategy of distillation 
networks that not only achieve 
state-of-the-art performance on SSL model distillation but also question our current understanding of
knowledge distillation.

\section{Related Works}
\label{sec:lit}
\textbf{Supervised knowledge distillation: } Bucilua \etal ~\cite{bucilua2006model} and Hinton \etal 
~\cite{hinton2015distilling} pioneered the use of knowledge 
distillation for compressing information. The methods used the teacher prediction logits as soft-labels in 
addition to the supervised label to regularize the student model. \cite{passalis2018learning} minimizes the divergence between
the student and teacher probability distributions. Several works 
(\cite{romero2014fitnets,komodakis2017paying,heo2019knowledge}) utilize intermediate teacher outputs in distillation.   
FitNets~\cite{romero2014fitnets} match both the final and intermediate teacher representations while \cite{komodakis2017paying} transfers knowledge from the attention maps of the teacher. RKD~\cite{park2019relational} 
transfers mutual relations instead of instance wise distillation.  
\cite{wang2020defense} proposes directly regressing the final teacher features with a modified loss function
that strictly matches the direction of the features but allows flexibility in terms of feature magnitude. 

\noindent \textbf{Self-supervised representation learning (SSL): }Earlier works on SSL 
(\cite{doersch2015unsupervised,gidaris2018unsupervised,caron2018deep,noroozi2016unsupervised,noroozi2017representation,pathak2016context})
learn effective representations by solving pretext tasks that do not require supervised labels. Recently, works 
based on contrastive learning (\cite{bachman2019learning,he2020momentum,chen2020improved,tian2019contrastive,chen2020simple,misra2020self,caron2020unsupervised}) have gained focus. In contrastive learning, the distances between representations
of positive pairs are minimized while those between negative pairs are maximized. The positive and negative pairs are 
generally constructed by utilizing multiple augmentations of each image. BYOL~\cite{grill2020bootstrap} is closer to 
knowledge distillation, where the distance between teacher and student representations are minimized. The inputs to the
two networks must be different augmentations of the same image and the teacher network is obtained as a moving average
of the student. Similar to our work, \cite{grill2020bootstrap} employs MLP head atop the student network to 
predict the teacher features. 

\noindent \textbf{Distillation of self-supervised models: }In \cite{noroozi2018boosting}, the student mimics 
the unsupervised 
cluster labels predicted by the teacher. CRD~\cite{tian2019contrastivedistill} maximizes a lower bound of the mutual
information between the teacher and student networks. However, it additionally uses supervised loss for optimization. 
CompRess~\cite{koohpayegani2020compress} and SEED~\cite{fang2021seed} are specifically designed for compressing 
self-supervised models. In both these works, student mimics the relative distances of teacher over a set of 
anchor points. Thus, they require maintaining large memory banks of anchor features and tuning temperature parameters.
As in regression, proposed prediction heads can also be used to improve CompRess and SEED.

\section{Knowledge Distillation}
\label{sec:method}

We first consider the supervised model distillation formulation proposed in \cite{hinton2015distilling}.
The teacher is trained on the task of object classification
from images. Let $X$ be the set of images, $Y$ the set of corresponding 
class labels and $c$ the total number of classes. Consider a teacher network \teach with
$f_t = \mathcal{T}(x)$, $f_t \in \mathbb{R}^c$ as the 
output vector (logits) corresponding to input image $x$. The predicted class probabilities can be obtained 
by applying softmax operation  $\sigma(.)$ atop the vector $f_t$.  
\vspace{-.1in}
\begin{equation}
    \hat{y}_t = \sigma(f_t; \tau_t) = \frac{e^{f_t/\tau_t}}{\sum_{i} e^{f_t^i/\tau_t}}
\end{equation}
where $f_t^i$ is the $i^{\textrm{th}}$ dimensional output of the feature vector and $\tau_t$ is the temperature parameter.
The teacher network \teach is trained using the image-label pairs in a supervised fashion
with standard cross-entropy loss. 
The trained teacher network is to be distilled to a student network. Once trained, the teacher 
network parameters are frozen during the distillation process.
Let \stud be the student network, $f_s = \mathcal{S}(x)$, $f_s \in \mathbb{R}^c$ the
feature vector corresponding to input image $x$ and $\hat{y}_s = \sigma(f_s; \tau_s)$ the predicted
class probability vector. Knowledge distillation loss is given by
\vspace{-.2in}
\begin{equation}
    \mathcal{L}_{\textrm{KD}}(\hat{y}_t, \hat{y}_s) = \sum_{j=1}^c \hat{y}_t^j \, \textrm{log}(\hat{y}_s^j)
\end{equation}
The student is trained using a combined objective function involving supervised cross entropy loss on student
features $\mathcal{L}_{\textrm{CE}}$ and distillation loss $\mathcal{L}_{\textrm{KD}}$:
\begin{equation}
    \mathcal{L} = \lambda \mathcal{L}_{\textrm{CE}} + (1 - \lambda) \tau_s^2 \mathcal{L}_{\textrm{KD}}
    \label{eq:loss_kd_total}
\end{equation}
where $\lambda$ is a hyperparameter that determines the relative importance of each loss term. 
Since $\tau_t$ is generally set
to 1, KD loss is multiplied by a factor of $\tau_s^2$ to match the scale of gradients from both loss terms.

\subsection{Distillation of Self-supervised Models}

The value of $\lambda$ in Eq.~\ref{eq:loss_kd_total} can be set to 0 if the class labels are not 
available during student distillation.
However, the formulation cannot be directly employed to distill from self-supervised teacher networks since the 
teacher outputs are latent representations and not logits or class probability vectors. 
Thus, to distill from such teachers, 
we simply regress the final feature vector of the teacher. Let $f_t = \mathcal{T}(x)$, $f_t \in \mathbb{R}^d$
and $f_s = \mathcal{S}(x)$, $f_s \in \mathbb{R}^m$. Since it is not necessary for the student and teacher 
representation dimensions to be the same, we use a linear projection of the student feature to match the 
dimensions. 
\begin{equation}
    f_s' = W^T f_s + b; \,\, W \in \mathbb{R}^m \times \mathbb{R}^d, \, b \in \mathbb{R}^d
    \label{eq:lin_proj}
\end{equation}
The distillation objective is then given by $\mathcal{L} = \mathcal{L}_{\textrm{reg}} = \textrm{d}(f_t, f_s')$
where $\textrm{d}(.)$ is a distance metric. Here, we consider squared Euclidean distance of $l_2$ normalized features as the metric.

\subsection{Prediction Heads for Regression Based Distillation}
For a more effective matching of the teacher latent space, we propose a non-linear prediction head $g(.)$ 
atop the student backbone network \stud in place of the linear projection in Eq.~\ref{eq:lin_proj}. 
During training, the student feature is then obtained as $f_s' = g(f_s)$ where $g(.)$ is modeled using 
a multi-layer perceptron (MLP). Each layer in $g(.)$ is given by a linear layer with bias followed by batch-normalization and a non-linear activation function (we use ReLU non-linearity in all our experiments).  
The number of such layers is a hyperparameter to be optimized. The dimension of the last layer output matches that of the
teacher. There is no non-linearity in the final layer to prevent constraining the output space of the student
network. During inference, the prediction head $g(.)$ is removed and the output of the student network is obtained as 
$f_s = \mathcal{S}(x)$ (refer Fig.~\ref{fig:arch_reg}). Thus, there is no change in the architecture or the number of parameters of the model
to be deployed. In our experiments, we demonstrate that the use of such MLP heads plays a crucial role in improving 
downstream performance. Surprisingly, we also observe that preserving the prediction heads during inference is not 
necessarily beneficial and might result in reduction in performance.  

\subsection{Multi-teacher Distillation}
The prediction heads are particularly beneficial in distillation from multiple teacher networks. Independent 
deep non-linear projections of the student backbone features can be employed during distillation to match
each of the teachers.
Let $f_t^k$ be the output vector of the $k^{th}$ teacher and $f_s^k = g^k(f_s)$ that of the corresponding 
student prediction head $g^k(.)$. The multi-teacher distillation objective for $K$ teachers is given by:
\vspace{-.05in}
\begin{equation}
    \mathcal{L} = \frac{1}{K} \sum_{k} \textrm{d}(f_t^k, f_s^k)
    \label{eq:loss_multi}
\end{equation}
The prediction heads are 
trained by the loss term from corresponding teachers while the backbone $S$ is trained using the summation in 
Eq.~\ref{eq:loss_multi}. 
\vspace{-.1in}

\section{Experiments}
\label{sec:exp}

We consider distillation of pretrained self-supervised models.
We consider four such methods for teacher networks - MoCo-v2~\cite{chen2020improved}, BYOL~\cite{grill2020bootstrap},
SwAV~\cite{caron2018deep} and SimCLR~\cite{chen2020simple}. We use the official publicly released models for all
the teacher networks (details in suppl.). We also use a ResNet-50 model trained with supervised labels (provided
by PyTorch in ~\cite{pytorchresnet}) as a teacher.
All teacher training and student distillation is performed on the train
set of ImageNet. We consider different teacher and student backbone network architectures.
For the prediction head, we experiment with linear, 2 and 4 layer MLPs. Let the dimension of 
the student backbone output be $m$ and that of teacher $d$. Similar to the prediction head in \cite{grill2020bootstrap}, the MLP dimensions are 
$(m, 2m, m, 2m, d)$.  

\noindent \textbf{Implementation details:} We use SGD optimizer with cosine scheduling of learning rate and momentum of 0.9. 
Initial learning rate is set to 0.05. As in \cite{koohpayegani2020compress} the networks are trained for 
130 epochs with batch size of 256. 
Cached teacher features are utilized for faster distillation in experiments with SimCLR, BYOL and SwAV teachers.
We publicly release the code\footnote{Code is available at \url{https://github.com/UCDvision/simreg}}.

\noindent \textbf{Datasets:} We primarily evaluate the performance of distilled networks on 
ImageNet~\cite{russakovsky2015imagenet} classification task.
Additionally, for transfer performance evaluation, we consider the following datasets: 
Food101~\cite{bossard2014food}, CIFAR10~\cite{krizhevsky2009learning}, CIFAR100~\cite{krizhevsky2009learning}, 
SUN397~\cite{sundatabase}, Cars~\cite{krause20133d}, Aircraft~\cite{maji2013fine}, DTD~\cite{cimpoi2014describing}, 
Pets~\cite{parkhi2012cats}, Caltech-101~\cite{fei2004learning} and Flowers~\cite{nilsback2008automated}. 
We train a single linear layer atop the frozen backbone network for transfer evaluation (refer suppl.)

\noindent \textbf{Metrics: }We use k-nearest neighbour (k-NN) and linear evaluation on all tasks. 
We also report mean squared error (MSE) between the student and teacher features over the test set. For k-NN 
evaluation, k=1 and 20 are considered and cosine similarity is used to calculate NNs. We employ FAISS~\cite{faiss} 
GPU library to perform fast k-NN
evaluation. For linear evaluation, a single linear layer is trained atop the features from the network to 
be evaluated. As in ~\cite{koohpayegani2020compress}, the inputs to the linear layer are normalized to unit 
$l_2$ norm and then each dimension is shifted and scaled to have unit mean and zero variance. The layer
is trained for 40 epochs using SGD with learning rate of 0.01 and momentum of 0.9.  

\begin{table}[]
    \centering
    \begin{tabular}{cccccc}
        \toprule
        Train and Inference Arch & 1-NN & 20-NN & Linear & MSE \\ \midrule
        MobileNet-v2+4L-MLP & \textbf{54.5} & \textbf{58.7} & \textbf{68.5} & \textbf{0.090}\\     
        MobileNet-v2+2L-MLP & 54.0 & 58.0 & 67.9 & 0.097\\     
        MobileNet-v2+Linear & 50.8 & 55.1 & 58.3 & 0.149\\     
        
        \bottomrule
    \end{tabular}
    \caption{\textbf{Role of MLP Heads.} We train three models with varying number of layers in prediction head and use the features from \textit{final} MLP layer of each prediction model for evaluation. 
    Deeper models more closely match the teacher (lower MSE) and achieve better classification performance (1 and 20 Nearest Neighbour and Linear evaluation).}
    \label{tab:mlpheads_depth}
\end{table}

\begin{table}[]
    \centering
    \begin{tabular}{cccc|ccc|ccc}
        \toprule
        Train & \multicolumn{9}{c}{Backbone(BB)+4L-MLP} \\
        Inference & \multicolumn{3}{c}{Backbone(BB)} & \multicolumn{3}{c}{BB+2L-MLP} & \multicolumn{3}{c}{BB+4L-MLP} \\
        \midrule
        Metric & 1-NN & Linear & MSE & 1-NN & Linear & MSE & 1-NN & Linear & MSE\\
        \midrule
        ResNet-18 & 55.3 & 65.7 & - & \textbf{56.0} & \textbf{66.4} & 1.99 & 53.4 & 65.2 & \textbf{0.1}\\
        \bottomrule
    \end{tabular}
    \caption{\textbf{Effect of MLP Heads on inference.} We train a single model with 4 layer MLP head and perform evaluation using features from different layers (pre-MLP, intermediate MLP layer and MLP output). Since the final layer outputs are trained to mimic the teacher, MSE with teacher features is lowest at 4L-MLP while that at 2L-MLP is extremely high (dimension of BB and teacher are different, hence MSE is not reported). However, features from backbone and intermediate layer (+2L-MLP) outperform those from final layer (+4L-MLP) on classification, contrary to the notion that features with lower MSE generalize better.}
    \label{tab:mlpheads}
\end{table}

\subsection{Baseline Approaches}
\textbf{Regression: }In addition to proposed MLP prediction head based regression (termed \textit{SimReg-MLP}), 
we consider two additional regression baseline methods proposed
in \cite{koohpayegani2020compress} termed `Regress' and `Regress-BN'. While Regress distills from unnormalized teacher 
features, Regress-BN uses batch-norm layer atop the final student and teacher features during distillation. Unlike 
Regress-MLP, both these  approaches use a linear prediction head.

\noindent \textbf{CompRess: } CompRess~\cite{koohpayegani2020compress} is designed to distill specifically from self-supervised models. 
Given a set of anchor points, the student is encouraged to have the same similarities with the anchors as that of the teacher. The
anchor point features can either be common features from a teacher memory bank (CompRess-1q) or features from individual 
memory banks for teacher and student (CompRess-2q). 
We additionally implement CompRess with our MLP 
prediction head, termed CompRess-1q-MLP and CompRess-2q-MLP. SEED~\cite{fang2021seed} proposes
similarity based distillation similar to ~\cite{koohpayegani2020compress} but uses pre-trained teacher
models with significantly lower number of training epochs and performance.
Further, it requires access to the projection heads used atop teacher networks used only during SSL training and not inference, 
These parameters are generally not publicly released, making the setting less replicable. 
Thus we provide comparisons with only CompRess~\cite{koohpayegani2020compress}.

\noindent \textbf{Contrastive Representation Distillation (CRD): } CRD~\cite{tian2019contrastivedistill} uses a 
contrastive formulation to bring corresponding teacher and 
student features closer while pushing apart those from unrelated pairs. While the paper considered a supervised setup and 
loss term utilizing labels, we use the formulation with just the contrastive loss as proposed in ~\cite{koohpayegani2020compress}.

\noindent \textbf{Cluster Classification (CC): } In CC~\cite{noroozi2018boosting}, the student predicts unsupervised labels 
obtained by clustering samples using teacher features. We report metrics for CC and CRD from ~\cite{koohpayegani2020compress}.

\section{Results}
\label{sec:res}
\begin{table}[]
    \centering
    \scalebox{0.92}{
    \begin{tabular}{lcc|cc|cc}
        \toprule
        Teacher & \multicolumn{2}{c}{MoCo-v2 ResNet-50} & \multicolumn{2}{c}{MoCo-v2 ResNet-50}  & \multicolumn{2}{c}{SimCLR ResNet-50x4}\\
        Student  & \multicolumn{2}{c}{MobileNet-v2} & \multicolumn{2}{c}{ResNet-18} & \multicolumn{2}{c}{ResNet-50} \\
        \midrule
        Method & 1-NN  & Linear & 1-NN  & Linear & 1-NN  & Linear \\
        \midrule
        Teacher*       & 57.3  & 70.8 & 57.3 & 70.8 & 64.5 & 75.6 \\
        Supervised*    & 64.9  & 71.9 & 63.0 & 69.8 & 71.4 & 76.2 \\
        \midrule
        Regress*    & 38.6  & 48.0 & 41.7 & 52.2 & -    & - \\
        Regress-BN* & 48.7  & 62.3 & 47.3 & 58.2 & -    & - \\
        CC~\cite{noroozi2018boosting}*         & 50.2     & 59.2 & 51.1 & 61.1 & 55.6 & 68.9 \\ 
        CRD~\cite{tian2019contrastivedistill}*        & 36.0     & 54.1 & 43.7 & 58.4 & -    & - \\
        CompRess-2q~\cite{koohpayegani2020compress}* & 54.4    & 63.0 & 53.4 & 61.7 & 63.0 & 71.0\\
        CompRess-1q~\cite{koohpayegani2020compress}* & 54.8    & 65.8 & 53.5 & 62.6 & 63.3 & 71.9\\
        \midrule
        CompRess-2q-4L-MLP & \textbf{56.3} & 67.4 & 54.4 & 64.0  & \textbf{62.5} & 73.5\\
        CompRess-1q-4L-MLP & 55.5 & 67.1  & \textbf{54.9} & 64.6 & 60.9 & 72.9 \\
        SimReg-4L-MLP  & 55.5 & \textbf{69.1} & 54.8 & \textbf{65.1} & 60.3 & \textbf{74.2}\\
        \bottomrule
    \end{tabular}}
    \caption{\textbf{Comparison of SSL distillation methods on ImageNet classification.} Our regression method with MLP head (SimReg-4L-MLP) is comparable to or better than the complex state-of-the-art approaches, especially on the linear evaluation metric. We also observe that CompRess-1q and 2q are improved when MLP heads are utilized. Interestingly, regression gets a significantly higher boost compared to CompRess upon addition of MLP layers. Note that the MLPs are used only during training and the inference network architecture remains the same for all approaches making the comparison fair. * metrics from CompRess~\cite{koohpayegani2020compress}.}
    \vspace{-0.1in}
    \label{tab:compare_distill}
\end{table}
\textbf{Role of Prediction Head:} A deeper prediction head results in a student with higher representational capacity and thus a model that 
better matches the teacher representations. Table~\ref{tab:mlpheads_depth} shows results for models with 
a common MobileNet-v2~\cite{sandler2018mobilenetv2} backbone and different prediction head architectures.
The prediction head is used during
both student training and evaluation. We observe that a deeper model has lower MSE with teacher features and 
better classification performance. However, a deeper model also implies greater inference time and 
memory requirements. The student architecture is fixed based on deployment needs and thus requirement of larger
model goes against the very essence of distillation. To analyze performance at different layers of the 
prediction head, we train a single ResNet-18~\cite{he2016deep} student with all intermediate dimensions of 
MLP equal to that of the output. Surprisingly, a 
model trained with MLP prediction head performs well on downstream task even when the prediction head is discarded 
during inference (Table~\ref{tab:mlpheads}). The performance using features from backbone network is slightly better than that from the final layer 
outputs whenever a MLP head is used (more results in suppl.). More importantly, this observation enables us to use 
deeper prediction heads for 
distillation in place of linear layers without any concerns about altering the student architecture or increasing 
inference time.\\

\noindent \textbf{Comparison with existing approaches:} In all the remaining experiments, we use SimReg-4L-MLP with the prediction head used only during distillation.
We compare the proposed regression method with other baselines and self-supervised distillation methods in 
tables~\ref{tab:compare_distill} and \ref{tab:imagenet_teach_ssl}. Surprisingly, our simple regression performs 
comparably or even outperforms the state-of-the-approaches on all settings and metrics. On linear evaluation, 
we outperform previous methods (without MLP) by \textbf{3.3}, \textbf{2.5} and \textbf{2.3} points respectively on MobileNet-v2, ResNet-18 and ResNet-50 students. Our observation generalizes 
to similarity based distillation too. Use of MLP prediction head also consistently improves the classification 
performance of both the CompRess variants (table~\ref{tab:compare_distill}). Note that the linear metrics of 
our student model on MobileNet-v2 and ResNet-50 are just 1.7 and 1.4 points below the corresponding teacher accuracies. Our ResNet-50 model distilled from SimCLR teacher outperforms a ResNet-50 model trained from scratch
using SimCLR (69.3\%~\cite{chen2020simple}) by \textbf{4.9} points.\\

\begin{table}[]
    \centering
    \begin{tabular}{lccc|ccc}
        \toprule
        Teacher & \multicolumn{3}{c}{BYOL ResNet-50}  & \multicolumn{3}{c}{SwAV ResNet-50} \\
        \midrule
        Method & 1-NN  & 20-NN & Linear & 1-NN  & 20-NN & Linear\\
        \midrule
        Teacher & 62.8 & 66.8 & 74.3 & 60.7 & 64.8 & 75.6 \\
        \midrule
        CompRess-2q-4L-MLP & 56.0 & 60.6 & 65.2 & 53.2 & 58.1 & 63.9 \\
        CompRess-1q-4L-MLP & 55.4 & 60.0 & 65.2 & 52.4 & 57.1 & 63.4 \\
        SimReg-4L-MLP  & \textbf{56.7} & \textbf{61.6} & \textbf{66.8} & \textbf{54.0} & \textbf{59.3} & \textbf{65.8} \\
        \bottomrule
    \end{tabular}
    \caption{\textbf{ImageNet Evaluation with different teacher networks.} We distill from two pretrained ResNet-50 SSL models, BYOL and SwAV to ResNet-18 students. When distilled from these stronger teacher networks, SimReg is significantly better than both CompRess variants on all metrics. Both SimReg and CompRess contain MLP head only during training.}
    \label{tab:imagenet_teach_ssl}
\end{table}

\noindent\textbf{Transfer Learning:} Since an important goal of self-supervised learning is to learn models that 
generalize well to new datasets and tasks, we evaluate the transfer learning performance of our distilled networks. 
The results in table~\ref{tab:full_transfer} suggest that the regression model transfers as 
well as or better than the state-of-the-approaches on most datasets. Among CompRess variants, 
CompRess-2q-MLP is generally better on ImageNet classification 
(table~\ref{tab:compare_distill}) but transfers poorly (table~\ref{tab:full_transfer}) compared to CompRess-1q-MLP. 
However, the same SimReg model performs comparably or outperforms them both in ImageNet and transfer tasks.  

\begin{table}[]
    \centering
    \scalebox{0.90}{
    \begin{tabular}{cc|ccc|ccc}
        \toprule
       Arch     & ResNet-50 & \multicolumn{3}{c}{MobileNet-v2} & \multicolumn{3}{c}{ResNet-18} \\
       Method   & Teacher & Comp-2q & Comp-1q & SimReg & Comp-2q & Comp-1q & SimReg \\ 
                &        & -4L-MLP   & -4L-MLP  & -4L-MLP & -4L-MLP   & -4L-MLP  & -4L-MLP \\ 
       \midrule
       Food     & 72.3 & 71.4  & 72.5   & \textbf{73.1} & 61.7 & \textbf{65.9} & 65.4\\ 
       CIFAR10  & 92.2 & 90.3  & 90.4   & \textbf{91.2} & 87.3 & \textbf{89.3} & 88.6\\ 
       CIFAR100 & 75.1 & 73.9  & 74.5   & \textbf{76.1} & 68.4 & \textbf{71.9} & 70.2\\ 
       SUN      & 60.2 & 58.0  & 58.1   & \textbf{59.4} & 54.3 & 56.0 & \textbf{57.1}\\ 
       Cars     & 50.8 & 60.3  & \textbf{63.1}   & 62.4 & 37.2 & \textbf{44.1} & 42.3\\ 
       Aircraft & 53.5 & 57.7  & \textbf{59.7}   & 58.7 & 42.3 & \textbf{47.8} & 45.8\\ 
       DTD      & 75.1 & 71.7  & 71.3   & \textbf{74.5} & 69.3 & \textbf{71.2} & 70.9\\ 
       Pets     & 83.6 & \textbf{86.7}  & 86.3   & 85.6 & 84.0 & \textbf{84.4} & 83.9\\ 
       Caltech  & 89.3 & 91.1  & 91.5   & \textbf{91.7} & 87.3 & \textbf{90.1} & 89.2\\ 
       Flowers  & 91.3 & 94.3  & \textbf{95.4}   & 95.1 & 86.4 & \textbf{91.3} & 90.9\\ \midrule
       Mean     & 74.3 & 75.5  & 76.3 & \textbf{76.8} & 67.8 & \textbf{71.2} & 70.4 \\ 
       \bottomrule
    \end{tabular}}
    \caption{\textbf{Transfer learning results on multiple classification tasks.} Since the teacher networks are self-supervised, generalization of learnt features to other datasets is important. SimReg is significantly better than CompRess-2q and comparable to CompRess-1q on most datasets. All methods employ 4 layer MLP heads only during distillation.}
    \label{tab:full_transfer}
\end{table}

In addition to transfer learning on different datasets on the classification task, we consider the task of object detection. 
We distill a ResNet-50 teacher to multiple ResNet-18 student networks with different MLP heads. The MLP heads are not part of 
the model fine-tuned for detection. Following~\cite{he2020momentum}, we use 
Faster-RCNN~\cite{ren2015faster} with \texttt{R18-C4} backbone architecture for the detection task. 
All methods are trained on the VOC \texttt{trainval07+12} and tested on \texttt{test07} subsets of the PASCAL 
VOC~\cite{everingham2010pascal} dataset using the code from \cite{pytorchmoco}. 
We report the standard AP$_{50}$, AP(COCO-style) and AP$_{75}$ metrics in table~\ref{tab:det_voc_r18}. 
Unlike in classification tasks, we find that the use of deeper MLP heads during distillation does not aid detection performance.
The performance of different distillation architectures is nearly identical on the detection task.\\

\begin{table}[]
    \centering
    \begin{tabular}{ccccc}
        \toprule
        Method & AP$_{50}$ & AP & AP$_{75}$ \\
        \midrule
        SimReg-4L-MLP      & 74.0 & 45.4 & 47.8\\
        SimReg-2L-MLP      & 74.2 & 45.5 & 47.4\\
        SimReg-Linear      & 73.6 & 45.1 & 47.9\\
        \bottomrule
    \end{tabular}
    \caption{\textbf{Transfer learning for object detection on Pascal VOC dataset.} Student models with different MLP head architectures are used to perform distillation on ImageNet dataset and the backbone with R18-C4 architecture is fine-tuned on PASCAL VOC. Unlike in classification tasks, the performance of different distillation architectures is nearly identical.}
    \label{tab:det_voc_r18}
\end{table}

\noindent \textbf{Effect of Data Augmentation:} As shown in SEED~\cite{fang2021seed} and CompRess~\cite{koohpayegani2020compress}, for a given student architecture, 
distillation from larger models trained using a particular method is better than directly training the student 
using the same method. Use of different and strong augmentations in contrastive SSL approaches has been shown to hurt
generalization performance~\cite{purushwalkam2020demystifying}. Here, we show that when distilling models, the best performance is obtained when the same 
augmented image with a weaker augmentation (details in suppl.) is used as input to both teacher and student networks (table~\ref{tab:aug_train}). This suggests that 
our method can be used to improve generalizability of SSL models. \\

\begin{table}[]
    \centering
    \begin{tabular}{ccccccc}
        \toprule
        Aug Type & \multicolumn{2}{c}{Aug Strength} &\multicolumn{3}{c}{ImageNet} & Transfer\\
        & Teacher & Student & 1-NN & 20-NN & Linear & Linear \\
        \midrule
        Same & Weak & Weak  & \textbf{54.8} & \textbf{59.9} & \textbf{65.1} & \textbf{70.4}\\
        Same & Strong & Strong & 53.4 & 59.0 & 64.3 & 70.3\\
        Different    & Weak & Weak & 54.7 & 59.8 & 64.6 & 70.0\\
        Different    & Weak & Strong   & 51.1 & 56.5 & 62.0 & 68.9\\
        Different    & Strong & Strong & 50.3 & 56.0 & 61.4 & 68.7\\
        \bottomrule
    \end{tabular}
    \caption{\textbf{Role of augmentation strength:} During distillation either the `same' augmented image or two `different' augmentations of a single image are used as inputs to the teacher and student networks. The augmentations strength is varied for both the settings. We find that the performance is best when the same image with weak augmentation is used. This is significant since using different and stronger augmentations improve classification performance of SSL models but decrease their generalizability~\cite{purushwalkam2020demystifying}.}
    \label{tab:aug_train}
\end{table}

\noindent \textbf{Multi-teacher Distillation:} We train a single student model from multiple teacher networks trained 
with different SSL methods. Regression with a 4 layer MLP head significantly outperforms one with linear prediction (table~\ref{tab:multi_teach}).\\

\begin{table}[]
    \centering
    \begin{tabular}{ccc|ccc|ccc}
        \toprule
        \multicolumn{3}{c}{Training} & \multicolumn{3}{c}{ResNet-18+Linear} & \multicolumn{3}{c}{ResNet-18+4L-MLP}\\
        \multicolumn{3}{c}{Inference}  & \multicolumn{3}{c}{ResNet-18} & \multicolumn{3}{c}{ResNet-18} \\
        \midrule
        MoCo-v2 & BYOL & SwAV & 1-NN  & 20-NN & Linear & 1-NN  & 20-NN & Linear\\
        \midrule
        \checkmark & \checkmark &  & 50.9 & 56.5 & 62.6 & 56.4 & 61.3 & 66.3\\
        \checkmark & & \checkmark  & 49.9 & 55.6 & 61.8 & 55.1 & 60.4 & 65.4\\
         & \checkmark & \checkmark & 52.0 & 57.7  & 64.8 & 56.5 & 61.4 & 67.0\\
        \checkmark & \checkmark & \checkmark & 51.1 & 56.7 & 63.2 & 56.5 & 61.6 & 66.9\\
        \bottomrule
    \end{tabular}
    \caption{\textbf{Multi-teacher distillation on ImageNet.} We train a single student model (ResNet-18) from multiple SSL teacher networks (ResNet-50) using a common backbone network and a separate prediction head for each teacher. Networks with 4 layer prediction heads can better match each of the teachers and thus vastly outperform those with a linear head on both k-NN and linear evaluation metrics.}
    \label{tab:multi_teach}
\end{table}

\noindent\textbf{Distillation of Supervised Teacher: }All the previous teacher networks were trained in a self-supervised manner. 
We additionally analyze the distillation from a teacher trained with supervision (table~\ref{tab:sup_dist}). 
Note that the distillation remains unsupervised and only the backbone CNN features of the teacher are regressed.
Similar to distillation of self-supervised teachers, we observe that the use of a deep MLP head during training significantly 
improves performance on the ImageNet classification task.

\begin{table}[]
    \centering
    \begin{tabular}{cccccc}
        \toprule
        Teacher & Student Arch & Prediction & 1-NN & 20-NN & Linear\\
                & (Inference) & Head (Train) \\
        \midrule
        \multirow{3}{*}{Supervised ResNet-50}& \multirow{3}{*}{\shortstack{MobileNet-v2 \\ Backbone}} & 4L-MLP & 63.77 & 67.87 & \textbf{73.5}\\
                                            & & 2L-MLP & \textbf{64.7} & \textbf{69.3} & \textbf{73.5}\\
                                            & & Linear & 55.4 & 62.0 & 67.5\\
        \bottomrule
    \end{tabular}
    \caption{\textbf{Distillation of supervised teacher.} Here, we analyze the role of MLP heads when distilling the features of a teacher trained using supervision. Note that the distillation remains unsupervised and only the backbone CNN features are regressed. Similar to distillation of SSL teachers, we observe that the use of a deep MLP head during training significantly improves classification performance on ImageNet classification task.}
    \label{tab:sup_dist}
\end{table}

\section{Conclusion}
\label{sec:con}

Distilling knowledge with deeper student networks leads to better downstream performance. 
We surprisingly find that intermediate layer outputs of a distilled student model 
have better performance compared to the final layer, though final layer is trained to mimic the
teacher representations. Thus, we use a prediction MLP head only for 
optimizing the distillation objective and achieve boosts in performance
with just the backbone network during inference.
We believe studying the reasoning for this effect is an interesting future work.
Our work also serves as an improved benchmark for future self-supervised distillation works. Additionally, 
we show that using the same 
weakly augmented image for both teacher and student aids distillation. 

\vspace{0.1in}
\noindent {\bf Acknowledgment:} 
This material is based upon work partially supported by the United States Air Force under Contract No. FA8750?19?C?0098, funding from SAP SE, and also NSF grant numbers 1845216 and 1920079. Any opinions, findings, and conclusions or recommendations expressed in this material are those of the authors and do not necessarily reflect the views of the United States Air Force, DARPA, or other funding agencies.

\bibliography{biblio}

\clearpage
\appendix
\setcounter{section}{0}
\setcounter{table}{0}
\setcounter{figure}{0}
\renewcommand{\thetable}{A\arabic{table}}
\renewcommand\thefigure{\arabic{figure}}
\renewcommand{\theHtable}{Supplement.\thetable}
\renewcommand{\theHfigure}{Supplement.\thefigure}

\section*{\huge Supplementary Material}

In this supplementary material, we present additional experimental results (Sec.~\ref{sec:add_exps}) and details on 
experiment settings and implementation (Sec.~\ref{sec:details}). Additional results include those on the role of 
MLP head during training (Sec.~\ref{subsec:mlp_role}) and self-distillation (Sec.~\ref{subsec:self_distill}). 
We publicly release the code\footnote{Code is available at \url{https://github.com/UCDvision/simreg}}.

\section{Additional Experimental Results}
\label{sec:add_exps}

\subsection{Role of MLP Head}
\label{subsec:mlp_role}

In tables 1 and 2 of main we analyze how the depth of MLP head during training and inference affects classification
performance. We present additional results here in table~\ref{tab:mlpheads_train_suppl} with different teacher 
and student network settings. The student networks are trained with different prediction head configurations. 
The evaluation is always performed using features from backbone network for a fair comparison. In addition to the 
self-supervised (SSL) teacher models used in the main paper, we consider a supervised teacher network. The teacher
is trained with cross-entropy loss using ground truth labels on the ImageNet dataset. As in SSL teachers, we use 
only the backbone network for distillation from a supervised teacher. Note that the supervised labels are absent
during student training. In both the supervised and self-supervised settings, the student with
4 layer MLP head consistently outperforms others on all metrics. \textbf{Compared to Linear head, 4L-MLP achieves 
5 (MoCo-v2, ResNet-18), 11.2 (MoCo-v2, MobileNet-v2) and 6 (Supervised, MobileNet-v2) percentage points improvement on 
linear evaluation}.

\begin{table}[]
    \centering
    \begin{tabular}{cccccc}
        \toprule
        Teacher & Student Arch & Prediction & 1-NN & 20-NN & Linear\\
                & (Inference) & Head (Train) \\
        \midrule
        \multirow{3}{*}{MoCo-v2 ResNet-50} & \multirow{3}{*}{\shortstack{ResNet-18 \\ Backbone}} & 4L-MLP & \textbf{54.8} & \textbf{59.9} & \textbf{65.1}\\
                                          & & 2L-MLP & 52.7 & 58.5 & 63.6\\
                                          & & Linear & 48.8 & 54.3 & 60.1\\
        \midrule
        \multirow{3}{*}{MoCo-v2 ResNet-50}& \multirow{3}{*}{\shortstack{MobileNet-v2 \\ Backbone}} & 4L-MLP & \textbf{55.46} & \textbf{59.73} & \textbf{69.1} \\
                                            & & 2L-MLP & 54.4 & 59.6 & 68.5\\
                                            & & Linear & 48.7 & 54.2 & 57.9\\
        \midrule
        \multirow{3}{*}{Supervised ResNet-50}& \multirow{3}{*}{\shortstack{MobileNet-v2 \\ Backbone}} & 4L-MLP & 63.77 & 67.87 & \textbf{73.5}\\
                                            & & 2L-MLP & \textbf{64.7} & \textbf{69.3} & \textbf{73.5}\\
                                            & & Linear & 55.4 & 62.0 & 67.5\\
        \bottomrule
    \end{tabular}
    \caption{\textbf{Effect of MLP Heads on ImageNet classification performance.} As in table 1 of main paper, we analyze the role of the prediction head used during training by varying the number of MLP layers. However, the evaluation here is performed using the features from the backbone network and the prediction head plays no role during inference. A linear prediction head corresponds to the architecture used in earlier works~\cite{koohpayegani2020compress}. We observe that a deeper prediction module during training results in substantial boosts in performance. This observation is consistent across different teacher networks (both SSl and supervised) and student architectures. \textbf{Compared to Linear head, 4L-MLP achieves 5(MoCo-v2, ResNet-18), 11.2(MoCo-v2, MobileNet-v2) and 6(Supervised, MobileNet-v2) percentage points improvement on linear evaluation}.}
    \label{tab:mlpheads_train_suppl}
\end{table}

\subsection{Self-distillation}
\label{subsec:self_distill}

In all the previous experiments, a larger teacher network is distilled to a shallower student. In 
self-distillation, we consider the same backbone architecture for both teacher and student. Similar
to other experiments, we use a prediction head (linear or MLP) atop student backbone during distillation
and remove it during evaluation. As we observe in table~\ref{tab:self_distill}, the student with a 
4 layer MLP head outperforms the teacher in both ImageNet classification and transfer tasks. The 
improvement in transfer performance is particularly significant (+4 percentage points) and might be 
attributed to the use of prediction head and weaker augmentations during distillation.

\begin{table}[]
    \centering
    \begin{tabular}{cccccc}
    \toprule
        Student Arch & Prediction Head & \multicolumn{3}{c}{ImageNet} & Transfer \\
        (Inference) & (Train) & 1-NN & 20-NN & Linear & Linear \\
        \midrule
         MoCo-v2 Teacher & - & 57.3 & 60.9 & 70.8 & 74.3\\
         \midrule
         ResNet-50 & 4L-MLP & \textbf{58.2} & \textbf{62.2} & \textbf{72.0} & \textbf{78.3}\\
         ResNet-50 & Linear & 56.4 & 60.6 & 69.7 & 71.8\\
     \bottomrule
    \end{tabular}
    \caption{\textbf{ImageNet Classification and transfer results for self-distillation with prediction head.} In self-distillation, the teacher and student backbone architectures are the same (ResNet-50). We use a MoCo-v2 pretrained teacher and train student networks with linear and 4 layer MLP heads. All evaluations are performed using backbone features. The student with MLP prediction head outperforms the teacher on both ImageNet classification and transfer tasks. A boost of 4 percentage points on average transfer accuracy suggests that the use of prediction head and weaker augmentations during distillation are beneficial in learning a good generalizable model.}
    \label{tab:self_distill}
\end{table}

\subsection{Comparison with CompRess without MLP}

In table 1 of the main paper, we observed that the use of MLP head during distillation benefits both the CompRess
variants on the ImageNet classification task. Here, we show that similar boost in CompRess performance can be achieved on 
transfer tasks when distilled with MLP head. We use the officially provided pretrained models for vanilla CompRess-2q ResNet-18 
and MobileNet-v2 architectures for our comparison and perform transfer analysis similar to that in table 5 of the main paper. 
Results in table~\ref{tab:transfer_compress_vanilla} demonstrate that performance of vanilla CompRess models are significantly
worse compared to both CompRess with MLP and proposed regression based distillation. Note that the MLP heads are not 
used during inference for fair comparison. 
\begin{table}[]
    \centering
    \scalebox{0.90}{
    \begin{tabular}{cccc|ccc}
        \toprule
      Arch     & \multicolumn{3}{c}{MobileNet-v2} & \multicolumn{3}{c}{ResNet-18} \\
      Method   & CompRess-2q & CompRess-2q & SimReg & CompRess-2q & CompRess-2q &  SimReg \\ 
                & plain       & -4L-MLP    & -4L-MLP &  plain & -4L-MLP    & -4L-MLP \\ 
      \midrule
      Food     & 61.4 & 71.4   & \textbf{73.1} & 57.6 & 61.7 & \textbf{65.4}\\ 
      CIFAR10  & 85.3 & 90.3   & \textbf{91.2} & 82.5 & 87.3 & \textbf{88.6}\\ 
      CIFAR100 & 65.1 & 73.9   & \textbf{76.1} & 62.5 & 68.4 & \textbf{70.2}\\ 
      SUN      & 53.9 & 58.0   & \textbf{59.4} & 52.2 & 54.3 & \textbf{57.1}\\ 
      Cars     & 35.0 & 60.3   & \textbf{62.4} & 30.0 & 37.2 & \textbf{42.3}\\ 
      Aircraft & 42.1 & 57.7   & \textbf{58.7} & 38.0 & 42.3 & \textbf{45.8}\\ 
      DTD      & 70.4 & 71.7   & \textbf{74.5} & 67.4 & 69.3 & \textbf{70.9}\\ 
      Pets     & 82.9 & \textbf{86.7}   & 85.6 & 81.6 & \textbf{84.0} & 83.9\\ 
      Caltech  & 85.6 & 91.1   & \textbf{91.7} & 85.3 & 87.3 & \textbf{89.2}\\ 
      Flowers  & 87.6 & 94.3   & \textbf{95.1} & 83.0 & 86.4 & \textbf{90.9}\\ 
      \bottomrule
    \end{tabular}}
    \caption{\textbf{Transfer learning performance of CompRess with and without MLP.} Since the teacher networks are self-supervised, generalization of learnt features to other datasets is important. Similar to ImageNet classification, CompRess with MLP significantly outperforms vanilla CompRess (CompRess-2q plain) on all datasets and metrics. MLP heads, if present, are only used during distillation and are not part of inference networks.}
    \label{tab:transfer_compress_vanilla}
\end{table}

\subsection{Results with Intermediate Layers of CNN}
In our results in table 2 of main paper, we analyze how the classification performance changes as we consider
features from the earlier layers of the prediction head. Here, we analyize results from various intermediate layers
including those from the CNN backbone. We train a single ResNet-18 student from a MoCo-v2 ResNet-50 teacher and 
perform k-NN evaluation using features from different layers. In table~\ref{tab:interm_layers}, Conv-1 refers to 
the output of the first convolutional layer while ResBlk-j refers to the output from the $j^{th}$ residual block.
The CNN features for evaluation are obtained by reducing their spatial dimension and then vectorizing. The spatial 
dimensions are reduced so that the feature lengths are roughly the same throughout the backbone for fair comparison. We 
observe that the performance increases as we go deeper into the backbone. The best performance is achieved at the
intermediate layer of prediction head and there is a small drop in accuracy at the final prediction layer.

\begin{table}[]
    \centering
    \scalebox{0.65}{
    \begin{tabular}{ccc|cc|cc|cc|cc|cc|cc}
        \toprule
        Eval Layer & \multicolumn{2}{c}{Conv-1} & \multicolumn{2}{c}{ResBlk-1} & \multicolumn{2}{c}{ResBlk-2} & \multicolumn{2}{c}{ResBlk-3} & \multicolumn{2}{c}{ResBlk-4} & \multicolumn{2}{c}{2L-MLP} & \multicolumn{2}{c}{4L-MLP}\\
        Student & 1-NN & 20-NN & 1-NN & 20-NN & 1-NN & 20-NN & 1-NN & 20-NN & 1-NN & 20-NN & 1-NN & 20-NN & 1-NN & 20-NN \\
        \midrule
        ResNet-18 & 6.2 & 7.2 & 16.0 & 18.0 & 21.8 & 24.3 & 33.4 & 37.4 & 55.3 & 60.2 & 56.0 & 60.9 & 53.4 & 57.6\\
        \bottomrule
    \end{tabular}}
    \caption{\textbf{ImageNet classification using intermediate features}. We consider a single student network with 4 layer MLP head and perform k-NN evaluation using features from various intermediate layer features from the network. For fair comparison, we match the dimensions from the intermediate convolutional features (Conv-1 and residual block features) to that of the final backbone feature (ResBlk-4) by reducing their spatial dimensions. As expected, performance improves as we use features from deeper layers of the CNN. This changes in the MLP head where a drop in accuracy at the very last layer of the prediction head is observed.}
    \label{tab:interm_layers}
\end{table}

\section{Implementation Details}
\label{sec:details}

\subsection{Teacher Networks}
We use teacher networks trained using four different self-supervised representation learning approaches - 
MoCo-v2~\cite{chen2020improved}, BYOL~\cite{grill2020bootstrap},
SwAV~\cite{caron2018deep} and SimCLR~\cite{chen2020simple}. We use the official and publicly available 
pre-trained weights for these networks with ResNet-50x4 architecture pretrained model 
for SimCLR teacher and ResNet-50 models for the remaining methods. MoCo-v2 and SwAV have been trained 
for 800 epochs and BYOL and SimCLR for 1000 epochs. For distillation with BYOL, SwAV and SimCLR teachers
we use cached features from the teacher. The cached features are obtained by passing the entire training
data through the teacher network once and storing the features. Random image augmentation as would be 
used in non-cached version is employed to generate the inputs for caching. 

\subsection{Image Augmentations}
We use two strategies for augmenting the input image during distillation - `weak' and `strong'. `Strong'
augmentation refers to the setting used in MoCo-v2~\cite{chen2020improved}. In both augmentation
settings, we apply a series of stochastic transformations on the input image. A random resized crop (scale
is in range [0.2, 1.]), random horizontal flip with probability 0.5 and normalization to channel-wise zero
mean and unit variance are common for both augmentation methods. In addition to these transformations,
`strong' augmentations use random color jittering (strength of 0.4 for brightness, contrast and saturation
and 0.1 for hue) with probability 0.8, random grayscaling with probability 0.2 and Gaussian blur (standard 
deviation chosen uniformly from [0, 1]). 

\subsection{Optimizer}
In all our distillation experiments, we use SGD optimizer with cosine scheduling of learning rate, momentum 
of 0.9 and weight decay of 0.0001. Initial learning rate is set to 0.05. The networks are trained for 130 
epochs with a batch size of 256 using PyTorch~\cite{pytorch} framework.  

\subsection{Evaluation Metrics}
We utilize k-NN and linear evaluation to evaluate classification performance on ImageNet and linear 
evaluation to evaluate transfer performance. For ImageNet linear evaluation, the inputs to the linear layer 
are normalized to unit $l_2$ norm and then each dimension is shifted and scaled to have unit mean and zero 
variance~\cite{koohpayegani2020compress}. The layer is trained for 40 epochs using SGD with initial learning 
rate of 0.01 and momentum of 0.9. The learning rate is scaled by 0.1 at epochs 15 and 30. For evaluation
of transfer performance, we use the optimizer settings from \cite{koohpayegani2021mean}. The shorter side of
the input image is resized to 256 and centre crop with length 224 is used. The input is channel-wise normalized
using the statistics from ImageNet dataset. We use LBFGS optimizer with parameters max\_iter=20 and 
history\_size=10. Learning rate and weight decay are optimized by performing a grid search using validation set.
The best model is obtained by retraining with optimal parameters on the combined train and validation set. 10
different log spaced values in [-3, 0] are used for learning rate while 9 log values in [-10, -2] are used for
weight decay.

\subsection{MLP Architecture}
For the proposed prediction head, we experiment with linear, 2 and 4 layer MLPs. Each MLP layer is composed of a
linear projection followed by 1D batch normalization and ReLU activation. Let the dimension of 
the student backbone output be $m$ and that of teacher $d$. For linear evaluation, a single layer with input 
and output dimensions of $(m, d)$ is used. For a 2 layer MLP, following \cite{grill2020bootstrap}, we use the 
dimensions $(m, 2m, d)$. We extend this to a 4 layer MLP with the following intermediate feature 
dimensions: $(m, 2m, m, 2m, d)$. Batch normalization and ReLU activation are not employed at the end of layer 2 
for 4 layer MLP head (equivalent to stacking two 2-layer MLP heads). For our ablation on the role of MLP head during 
inference (table 2 in main
paper), we compare the performance of our method at different layers of the MLP head from a single 
trained network. For fair comparison, we require all the intermediate dimensions to be same as that of the 
output. Thus, for this experiment alone, we use an MLP such that the feature dimensions are 
$(m, d, d, d, d)$. The output dimension ($m$) for ResNet-18, ResNet-50 and MobileNet-v2 are 512, 2048 and 1280
respectively. The teacher output dimensions are 2048 and 8192 respectively for ResNet-50 and ResNet-50x4 
architectures. From table 2 (network with MLP feature dimensions (512, 2048, 2048, 2048, 2048)) and table 4
(network with MLP feature dimensions (512, 1024, 512, 1024, 2048)) results, we observe that higher MLP feature
dimensions might help further boost performance (65.7 vs 65.1 on ImageNet linear). More ablations on this are
necessary to optimize the MLP architecture.

\end{document}